\pdfoutput=1

\documentclass[11pt]{article}

\usepackage[table]{xcolor}

\usepackage[]{acl}

\usepackage{enumitem}
\setitemize{noitemsep,topsep=0pt,parsep=0pt,partopsep=0pt}

\usepackage{times}
\usepackage{latexsym}

\usepackage[T1]{fontenc}

\usepackage[utf8]{inputenc}

\usepackage{microtype}

\usepackage{inconsolata}

\usepackage{microtype}
\usepackage{graphicx}
\usepackage{subfigure}
\usepackage{booktabs} 

\usepackage{amsmath}
\usepackage{amssymb}
\usepackage{mathtools}
\usepackage{amsthm}

\usepackage{url}

\definecolor{light-gray}{gray}{0.98}
\definecolor{med-gray}{gray}{0.95}

\title{Dynamic Masking Rate Schedules for MLM Pretraining}

\author{Zachary Ankner $^{1,2}$ $\,\,$ Naomi Saphra$^{3}$ $\,\,$ Davis Blalock$^{1}$ \\ \textbf{Jonathan Frankle}$^{1}$ $\,\,$ \textbf{Matthew Leavitt}$^4$\\
$^1$MosaicML $\,\,\,$ $^2$Massachusetts Institute of Technology $\,\,\,$ \\ $^3$Harvard University $\,\,\,$ $\,\,\,$ $^4$DatologyAI$\,\,\,$
}

\begin{document}

\maketitle

\begin{abstract}
Most works on transformers trained with the Masked Language Modeling (MLM) objective use the original BERT model's fixed masking rate of 15\%.
We propose to instead dynamically schedule the masking rate throughout training.
We find that linearly decreasing the masking rate over the course of pretraining improves average GLUE accuracy by up to 0.46\% and 0.25\% in \texttt{BERT-base} and \texttt{BERT-large}, respectively, compared to fixed rate baselines.
These gains come from exposure to both high and low masking rate regimes, providing benefits from both settings. 
Our results demonstrate that masking rate scheduling is a simple way to improve the quality of masked language models, achieving up to a 1.89x speedup in pretraining for \texttt{BERT-base} as well as a Pareto improvement for \texttt{BERT-large}.
\end{abstract}

\newcommand\blfootnote[1]{%
  \begingroup
  \renewcommand\thefootnote{}\footnote{#1}%
  \addtocounter{footnote}{-1}%
  \endgroup
}

\blfootnote{Correspondence to \href{ankner@mit.edu}{ankner@mit.edu}.}

\section{Introduction}

BERT \cite{bert-delvin} is a popular encoder-only Transformer \cite{attention-vasawani} architecture that is pretrained using a Cloze-inspired~\cite{taylor_cloze_1953} masked language modeling (MLM) objective.
During MLM training, we mask out a subset of the input tokens and train the model to reconstruct the missing tokens. The proportion of tokens to be masked out is determined by the \emph{masking rate} hyperparameter.  

Most practitioners use a fixed masking rate of 0.15 \citep{bert-delvin}, but \citet{bert-rate-wettig} found that the standard 15\% masking rate is sub-optimal for a variety of model settings and recommended a higher rate. We build on their work by studying the impact of dynamically scheduled masking rates.

Hyperparameter scheduling---i.e., changing the learning rate, dropout rate, batch size, sequence length, etc., during training---is a common practice in deep learning~\cite{sgdr-loshchilov, cyclical-smith, ulmft-howard, curriculum-dropout-morerio, bs-sched-smith, sls-li}. Masking rate is a good candidate for hyperparameter scheduling for a number of reasons. 
First, a high masking rate, like a high dropout rate, directly reduces the amount of feature information available during a training step. This information removal may smooth the loss landscape, which permits simulated annealing if performed earlier in training.
Furthermore, a higher masking rate adds training signal, as loss is computed for a larger portion of tokens, similar to a larger sequence length or batch size.
We therefore study whether scheduling the masking rate during training could lead to model quality improvements, as scheduling these other hyperparameters does.

We present a series of experiments to assess the effects of masking rate scheduling on the quality of \texttt{BERT-base} \cite{bert-delvin}. We evaluate our masking rate scheduled models on MLM loss and downstream tasks. Our contributions are:
\begin{itemize}
    \item We introduce a method of masking rate scheduling\footnote{After submitting this work, we were made aware of recent work \citep{time-variant-masking-yang} that also applies dynamic masking rates to MLM pretraining. Our method for scheduling masking rates differs slightly but our analysis of the technique substantially differs by focusing on understanding how scheduling improves MLM performance. We discuss these differences in Section~\ref{sec:related-work}.} for improving MLM pretraining (Section~\ref{sec:best-perf}), and find that performance improves only when starting at a higher ratio and 
    decaying it (Section~\ref{sec:higher-init-masking}).
    \item We show that the improvement from scheduling the masking rate is a Pareto improvement over fixed masking rates~(Section~\ref{subsec:efficiency}, Appendix~\ref{sec:appendix-large-pareto}), and that our method transfers to other pretraining objectives (Appendix \ref{sec:appendix-rts}).
    \item We find that dynamic scheduling attains both the improved linguistic performance of a lower masking rate (Section~\ref{sec:grammar}) and improved language modeling of a higher masking rate~(Section~\ref{sec:mlm-perf}).
\end{itemize}

\begin{table*}[t]
\begin{center}
\begin{small}
\begin{sc}
\begin{tabular}{lccccccccc}
\toprule
 Schedule & MNLI-m/mm & QNLI & QQP & RTE & SST-2 & MRPC & CoLA & STS-B & Avg \\
\midrule

\rowcolor{light-gray} \textit{BERT-base} & & & & & & & & & \\

\rowcolor{light-gray} \hspace{1mm} constant-0.15	& 84.3/84.71	& 90.38	& \textbf{88.31}	& \textbf{76.65}	& \textbf{92.91}	& \textbf{91.94}	& 55.89	& 89.38	& 83.83		 \\

\rowcolor{light-gray} \hspace{1mm} constant-0.3	& 84.5/84.83	& \textbf{90.82}	& \textbf{88.31}	& \textbf{76.56}	& \textbf{92.79}	& \textbf{92.18}	& 57.24	& \textbf{89.85}	& 84.12		 \\

\rowcolor{light-gray} \hspace{1mm} linear-0.3-0.15 \scriptsize(Ours)	& \textbf{84.61}/\textbf{85.13}	& \textbf{90.89}	& \textbf{88.34}	& \textbf{76.25}	& \textbf{92.71}	& \textbf{91.87}	& \textbf{58.96}	& \textbf{89.87}	& \textbf{84.29}		 \\

\midrule 

\rowcolor{med-gray} \textit{BERT-large} & & & & & & & & & \\

\rowcolor{med-gray} \hspace{1mm} constant-0.4	& 87.43/87.68	& 93.03	& \textbf{88.84}	& \textbf{83.25}	& \textbf{94.48}	& \textbf{93.64}	& \textbf{63.53}	& 90.82	& 86.97		 \\

\rowcolor{med-gray} \hspace{1mm} linear-0.4-0.25	& \textbf{87.69}/\textbf{87.9}	& \textbf{93.33}	& \textbf{89.23}	& \textbf{83.14}	& \textbf{94.59}	& \textbf{93.86}	& \textbf{64.07}	& \textbf{91.21}	& \textbf{87.22}		 \\

\bottomrule
\end{tabular}
\end{sc}
\end{small}
\end{center}
\caption{Downstream performance for different masking rate schedules. For each model we report the average accuracy for each task in GLUE. Bold indicates no significant difference from best-performing schedule, \emph{P > 0.05}, t-test.}
\label{tab:best-perf}
\end{table*}

\section{Methods}
We perform typical MLM pretraining, with the key difference that a scheduler sets the masking rate dynamically.

\subsection{Masked language modeling}
\label{sec:mlm-obj}
An MLM objective trains a language model to reconstruct tokens that have been masked out from an input sequence.
Let $x \sim \mathcal{X}$ be the input sequence, and $p_{\text{mask}}$ be the probability with which tokens are masked from the model, i.e., the masking rate.
A mask $\mathcal{M}=\{m_1,...,m_k\}$ is defined as the indices of the tokens to be masked, where the probability of a given token index being included in the mask is a Bernoulli random variable with parameter $p_{\text{mask}}$.
Following \citet{bert-delvin}, we replace 80\% of the masked tokens with a \verb=[MASK]= token, substitute 10\% with another random token, and leave 10\% unchanged.
The training objective is defined as:
\begin{equation}
\mathcal{L}(x) = \frac{1}{|\mathcal{M}|}\sum_{i \in \mathcal{M}} \mspace{4mu} \log \mspace{4mu} p(x_{m_i}|x_{-\mathcal{M}})
\end{equation}

\subsection{Schedulers}

Let $\mathcal{T}_{\texttt{total}}$ be the total number of steps the model takes during training and $t$ be the current step.
Let $p_i$ and $p_f$ be the initial and final masking rate respectively.
For each step, we set the masking rate $p_{\text{mask}, t}$ according to the following schedules. 
We test several nonlinear schedules as well, but find no consistent advantage over the simpler linear schedule (Appendix~\ref{sec:appendix-other-sched}).

\paragraph{Constant scheduling.}
Constant scheduling, which we call \mbox{\texttt{constant-\{$p_\text{mask}$\}}}, is the standard approach to setting the masking rate for MLM pretraining (typically $p_{\text{mask}} = 0.15$) where the same masking rate is used throughout all of training.
The masking rate is set as:

\[p_{\text{mask}, t}=p_i=p_f\]

\paragraph{Linear scheduling.}
In the linear schedule \texttt{linear-\{$p_i$\}-\{$p_f$\}}, the masking rate is set to a linear interpolation between the initial and final masking rate:
\[p_{\text{mask}, t}=p_i + \frac{t}{\mathcal{T}_{\texttt{total}}}*(p_f - p_i)\]
\\[-6mm]

\section{Experiments and Results}

In this section, we evaluate the performance of masking rate scheduling on a collection of downstream tasks and determine why our schedule is successful. 

We pretrain all models on the Colossal Cleaned Common Crawl (C4) dataset \cite{c4-raffel}, and then fine-tune and evaluate on the GLUE benchmark \cite{glue-wang}.
We use \texttt{BERT-base} and \texttt{BERT-large} models as implemented in HuggingFace \cite{huggingface-wolf}, and train models with the Composer library \cite{tang_composer_2022}.
We list further details of our experimental setup in Appendix~\ref{sec:appendix-training}.

\subsection{Improvement in downstream tasks}
\label{sec:best-perf}

We first examine the effects of the best linear schedule on downstream performance on GLUE (Table~\ref{tab:best-perf}).
We focus on comparing between \texttt{linear-0.3-0.15} and \texttt{constant-0.3-0.3} for \texttt{BERT-base}, and between \texttt{linear-0.4-0.25} and \texttt{constant-0.4-0.4} for \texttt{BERT-large}. These settings provide the best-performing linear and constant schedules, respectively. (Results for other schedule hyperparameters are in Appendix~\ref{sec:appendix-sched-sweep}.)
For \texttt{BERT-base}, we find that \texttt{linear-0.3-0.15} improves performance over the baseline on 3 of the 8 GLUE tasks and achieves parity on all other tasks, leading to an average GLUE accuracy of 84.29\%, a statistically significant improvement over the \texttt{constant-0.3-0.3} baseline of 84.12\%.
For \texttt{BERT-large} we find that \texttt{linear-0.4-0.25} improves performance over the baseline on 4 of the 8 GLUE tasks and achieves parity on all other tasks, leading to an average GLUE accuracy of 87.22\%, a statistically significant improvement over the \texttt{constant-0.4-0.4} baseline of 86.97\%.
These results show that scheduling the masking rate during pretraining produces higher-quality models for downstream tasks.

\subsection{Improvement in training efficiency}
\label{subsec:efficiency}
\begin{figure}[ht]
    \begin{center}
    \centerline{\includegraphics[width=\columnwidth]{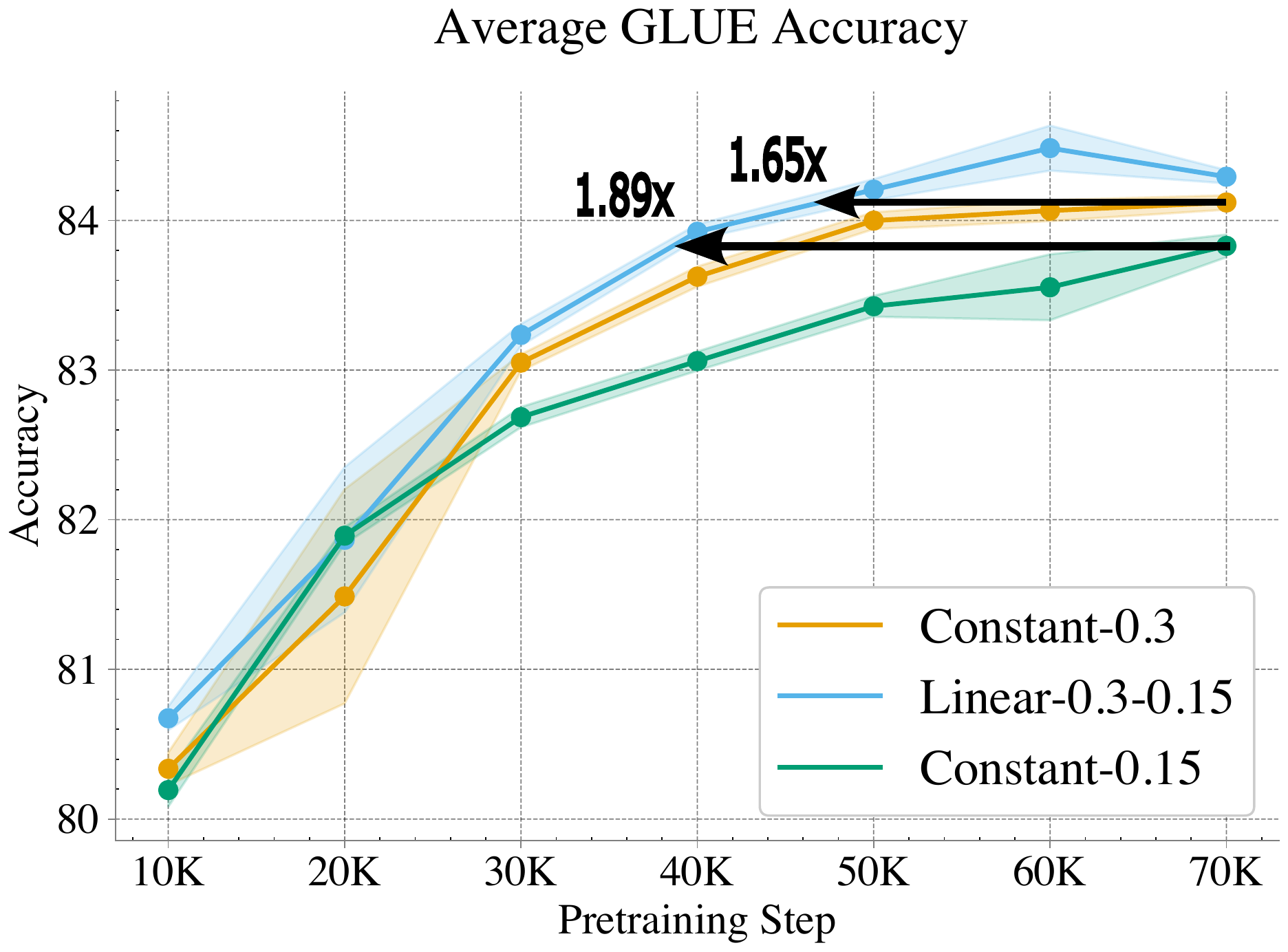}}
        \caption{Average GLUE accuracy evaluated over the course of pretraining for \texttt{BERT-base}. The horizontal lines correspond to the difference in steps required for \texttt{linear-0.3-0.15} to achieve the best constant schedule performance.}
        \label{fig:speed-up}
    \end{center}
\end{figure}

In addition to improving final model quality, pretraining with masking rate scheduling is more efficient in wall clock time.
For \texttt{BERT-base}, linear scheduling matches the mean GLUE score of the best $\texttt{constant-0.15}$ checkpoint in 37K steps and matches the best $\texttt{constant-0.3}$ checkpoint in 42K steps, which correspond to speedups of 1.89x and 1.65x, respectively.
Furthermore, $\texttt{linear-0.3-0.15}$ is a Pareto improvement over both constant baselines; for each pretraining step evaluated, $\texttt{linear-0.3-0.15}$ matches or exceeds the baseline with no increase in training time~(Figure~\ref{fig:speed-up}).
For \texttt{BERT-large}, \texttt{linear-0.4-0.25} is also a Pareto improvement over \texttt{constant-0.4} (Appendix~\ref{sec:appendix-large-pareto}). 
Appendix~\ref{sec:appendix-speed-up} contains further details on evaluating model speedups.

\subsection{High to low, not low to high}
\label{sec:higher-init-masking}

To better understand how masking rate scheduling affects training dynamics, we investigate whether the scheduler must always gradually decrease the masking rate, in line with an interpretation based on simulated annealing \citep{kirkpatrick1983optimization}. If we find that either decreasing or increasing lead to similar improvements, then we instead would attribute the success of our method to just the range of masking rates covered.
We find that the reversed schedule \texttt{linear-0.15-0.3} performs significantly worse than the decreasing schedule \texttt{linear-0.3-0.15} on GLUE for \texttt{BERT-base}, and in fact has performance comparable to the \texttt{constant-0.15} baseline (Table~\ref{tab:glue-increasing-decreasing}).

\begin{table}[t]
\begin{center}
\begin{small}
\begin{sc}
\begin{tabular}{lc}
\toprule
Schedule & Avg GLUE Accuracy \\
\midrule
constant-0.15 & 83.83 \\
linear-0.15-0.3 & 83.71 \\
linear-0.3-0.15 & \textbf{84.29}\\
\bottomrule
\end{tabular}
\end{sc}
\end{small}
\end{center}
\caption{Average GLUE accuracy for increasing/decreasing schedules with the same range of masking rates. Bold indicates no significant difference from the highest-performing schedule, \emph{P > 0.05}, t-test.}
\label{tab:glue-increasing-decreasing}
\end{table}

\subsection{Masking and loss are both necessary for improved performance}
\label{sec:subset-mask}
Is the added signal from a dynamic masking rate necessary, or does the removal of information from the inputs determine the majority of the gain? Here, we distinguish two possible sources of benefit from our schedule: benefits from smoothing the loss surface; and benefits from adding training examples by increasing the number of masked words to predict.
To test whether the latter is necessary, we pretrain a \texttt{BERT-base} model linearly scheduling the masking rate from 30\% to 15\%, but we only compute the loss on a subset of the masked tokens such that the loss is defined over 15\% of the input tokens (referenced as \texttt{subset-linear-0.3-0.15}).
We find that \texttt{subset-linear-0.3-0.15} under-performs both \texttt{linear-0.3-0.15} and \texttt{constant-0.15} (Table~\ref{tab:structure-only}).
This result suggests that obfuscating the input sequence according to a dynamic masking rate does not by itself improve modeling performance, and thus the increased signal is also necessary.

\begin{table}[t]
\begin{center}
\begin{small}
\begin{sc}
\begin{tabular}{lc}
\toprule
Schedule & Avg GLUE Accuracy \\
\midrule
constant-0.15	&  83.83		 \\
subset-linear-0.3-0.15	&  83.71		 \\
linear-0.3-0.15		& \textbf{84.29}		 \\
\bottomrule
\end{tabular}
\end{sc}
\end{small}
\end{center}
\caption{Average GLUE score for scheduling masking rate while holding constant the number of tokens used in training. Bold results show no significant difference  (t-tested $p$ < 0.05) from the highest-performing schedule.}
\label{tab:structure-only}
\end{table}

\subsection{Improvement in grammar capabilities}
\label{sec:grammar}

In order to better understand scheduling's effects on the linguistic capabilities of MLMs, we evaluated our models on the BLiMP benchmark~\cite{blimp-warstadt}; this benchmark tests understanding of syntax, morphology, and semantics.

We find the average BLiMP accuracy of \texttt{linear-0.3-0.15} significantly improves over \texttt{constant-0.3} and matches $\texttt{constant-0.15}$ (Table~\ref{tab:blimp-avg}). These results suggest that a dynamic schedule enables the linguistic capabilities of a lower masking rate.

\begin{table}[t]
\begin{center}
\begin{small}
\begin{sc}
\begin{tabular}{lc}
\toprule
Schedule & Avg BLiMP Accuracy \\
\midrule
linear-0.3-0.15 & \textbf{82.70}\\
constant-0.15 & \textbf{82.44}\\
constant-0.3 & 82.13\\
\bottomrule
\end{tabular}
\end{sc}
\end{small}
\end{center}
\caption{Average accuracy across BLiMP tasks. Bold indicates mean + standard error matches best average.}
\label{tab:blimp-avg}
\end{table}

\subsection{Improvement in the pretraining objective}
\label{sec:mlm-perf}

How does a decreasing schedule affect a model's language modeling ability?
When evaluating models at a 15\% masking rate, we find that \texttt{linear-0.3-0.15} and \texttt{constant-0.3} have the same average MLM loss of 1.56.
However, \texttt{constant-0.15} performs significantly worse, with a best MLM loss of 1.59.

Although scheduling only temporarily sets the masking ratio close to 30\%, scheduled models match the superior language modeling capabilities of 30\% masking throughout the entire pretraining duration.

\section{Related work}
\label{sec:related-work}
\paragraph{Masked Language Modeling}

Since ELMo \citep{peters-etal-2018-deep}, self-supervised pretraining has become the dominant paradigm for many NLP tasks, and BERT has been established as a basic standard for transfer learning.
Many works have changed the BERT model architecture while retaining the original MLM objective, including the 15\% constant masking rate \cite{roberta-liu, albert-lan, bigbird-zaheer, deberta-he}. Other encoder-only models have modified the MLM objective itself to mask out spans of tokens instead of individual tokens \cite{span-bert-joshi, ernie-zhang, pmi-masking-levine}.
We note that both architectural changes and span masking are compatible with our masking rate scheduling.

ELECTRA~\cite{electra-clark} proposes an alternate denoising objective to masking; using a separate ``generator'' encoder language model, they replace a subset of tokens in the input sequence.
While the gradual improvement of the generator may implicitly parallel a masking rate schedule, explicit scheduling may still be beneficial since accuracy can be sensitive to masking rate (Appendix~\ref{sec:appendix-other-sched}).
Additionally, the generator is trained using an MLM objective, and as such could benefit from masking rate scheduling.

There has also been previous work exploring whether the standard 15\% masking rate is optimal.
\citet{bert-rate-wettig} empirically investigate the optimal fixed masking rate and demonstrate that for larger BERT models higher masking rates are more performant.

Most closely related to our method is \citet{time-variant-masking-yang}, which also examines dynamic masking rates for MLM pretraining.
Although there is significant overlap in the proposed methodologies, their work sets the final masking rate to be close to 0\%, while we found that maintaining a higher final masking rate of 15\% was necessary for performance improvements.
Additionally, our analysis differs significantly from theirs.
While both their work and ours evaluate downstream performance improvements, \citet{time-variant-masking-yang} also investigates how dynamic masking rates affect performance when the training duration is extended and study nonrandom token masks.
Our analysis, by contrast, focuses on why masking rate scheduling improves performance. To this end, we investigating whether dynamic masking rates must follow a decaying scheduling (Section~\ref{sec:higher-init-masking}), whether the observed gains are due to the additional training signal or the added noise (Section~\ref{sec:subset-mask}), the impact of differing masking rate schedules on grammatical capabilities (Section~\ref{sec:grammar}), and the impact of dynamic masking rates on the pre-training objective itself (Section~\ref{sec:mlm-perf}).

\paragraph{Hyperparameter scheduling} Although learning rate is the most commonly-scheduled hyperparameter~\cite{sgdr-loshchilov,cyclical-smith, ulmft-howard}, other hyperparameter schedules are common. Our approach is also not the first to schedule a hyperparameter that removes information content from the model; prior work has suggested scheduling dropout~\cite{curriculum-dropout-morerio, drophead-zhou} and input resolution~\cite{progressive-resize-fastai}.
Scheduling has also been applied to hyperparameters that control the training signal to the model such as batch size~\cite{bs-sched-smith} and sequence length~\cite{sls-li}. Masking rate combines both of these properties, making it a particularly good candidate for scheduling.

\section{Discussion and Conclusions}

In addition to our method's improvement on the average final downstream performance, we find that scheduling is a Pareto improvement for \emph{all examined pretraining durations} over the typical constant masking rate baselines on GLUE.
Our analysis suggests that this benefit comes from the combined advantages of higher and lower masking rates.
We also demonstrate that our approach generalizes to other pretraining objectives (Appendix~\ref{sec:appendix-rts}).

Our method of beginning with a larger masking ratio and decaying, which we found necessary (Section~\ref{sec:higher-init-masking}), parallels the motivation behind \textit{simulated annealing} \cite{kirkpatrick1983optimization}.
Simulated annealing is a general method for avoiding local minima by smoothing the loss surface early in training through the addition of noise early in training.
However, we found that the increasing noise early in training is not the only source of advantage. We also benefit from increasing the signal by predicting more masked tokens (Section~\ref{sec:subset-mask}).

Overall, our work demonstrates that masking rate scheduling is a simple and reliable way to improve the quality and efficiency of MLM pretraining.

\section*{Limitations}
In this work, we restrict ourselves to English-only pretraining and finetuning. For other languages with free word order, there may be less information about the overall sentence structure when masking at a higher rate because the position of a word provides less information. 
As such our technique may not generalize or be suitable for other languages.

Additionally, we only investigate masking rate scheduling in the encoder setting. Further applying our method to encoder-decoder settings where the model is partially trained with a reconstruction loss, such as T5, is a direction for future research. 

Finally, we only evaluate models on the GLUE benchmark. While our evaluation is in line with previous work, a more comprehensive set of tasks could provide a better evaluation.

\section*{Acknowledgments}

While performing this work, Naomi Saphra was employed by New York University and Matthew Leavitt was employed by MosaicML.
This work was supported by Hyundai Motor Company (under the project Uncertainty in Neural Sequence Modeling) and the Samsung Advanced Institute of Technology (under the project Next Generation Deep Learning: From Pattern Recognition to AI). 

\bibliography{references}

\appendix

\section{Training Details}
\label{sec:appendix-training}
\paragraph{Modeling details.}
We use a \texttt{BERT-base} and \texttt{BERT-large} model as implemented in HuggingFace \cite{huggingface-wolf}, which have 110 million and 345 million parameters respectively.
To manage the training of models we use the Composer library \cite{tang_composer_2022}.
All training is conducted on 8 NVIDIA A100 GPUs.
\texttt{BERT-base} and \texttt{BERT-large} take approximately 10 hours and 24 hours to train respectively.

\paragraph{Pretraining.}
For our \texttt{BERT-base} experiments, we perform 3 trials of MLM pretraining on a 275 million document subset of the Colossal Cleaned Common Crawl (C4) dataset \cite{c4-raffel}.
For \texttt{BERT-large} experiments, we perform 2 trials of MLM pretraining for 2 epochs of the C4 dataset.
For all models, following a learning rate warm-up period of 6\% of the total training duration, we linearly schedule the learning rate from 5e-4 to 1e-5.
We use the AdamW optimizer \cite{adamw-Loshchilov} with parameters $\beta_1=0.9$, $\beta_2=0.98$, $\epsilon=\texttt{1e-6}$, and a decoupled weight decay of $\texttt{1e-5}$. 
All models are trained using a sequence length of~\texttt{128} and a batch size of \texttt{4096}.

\paragraph{Downstream evaluation.}
We fine-tune and evaluate all models on the GLUE benchmark \cite{glue-wang} which is composed of a variety of tasks evaluating different natural language tasks. All fine-tuning results are repeated for 5 trials for each pretraining trial.

\section{Significance testing}
For a given task, to determine whether a masking rate schedule has performance comparable to the masking rate schedule with the best mean performance across seeds, we compute a one-sided t-test of the hypothesis "Schedule $X$ performs worse than schedule $Y$", where $X$ is the schedule being compared and $Y$ is the schedule with the best mean performance.
Since we are computing multiple pair-wise t-tests, we correct the pairwise t-tests using the Hochberg step-up procedure~\cite{hochberg-test}.
If the corrected P-value is less than 0.05 we reject the null hypothesis and conclude that the schedule with the greater mean performance significantly outperforms the alternative schedule.

\section{Sweeping Schedule Hyperparameters}
\label{sec:appendix-sched-sweep}

\begin{table*}[t]
\begin{center}
\begin{small}
\begin{sc}
\begin{tabular}{lccccccccc}
\toprule
Schedule & MNLI-m/mm & QNLI & QQP & RTE & SST-2 & MRPC & CoLA & STS-B & Avg \\
\midrule

\rowcolor{light-gray} \textit{Constant} & & & & & & & & & \\

\rowcolor{light-gray} \hspace{1mm} constant-0.15	& 84.3/84.71	& 90.38	& \textbf{88.31}	& 76.65	& \textbf{92.91}	& \textbf{91.94}	& 55.89	& 89.38	& 83.83		 \\

\rowcolor{light-gray} \hspace{1mm} constant-0.2	& \textbf{84.46}/\textbf{84.95}	& 90.64	& \textbf{88.24}	& 76.73	& 92.59	& \textbf{91.63}	& \textbf{56.45}	& 89.6	& 83.92		 \\

\rowcolor{light-gray} \hspace{1mm} constant-0.25	& 84.28/84.79	& 90.61	& \textbf{88.3}	& 76.27	& 92.54	& \textbf{92.06}	& \textbf{56.74}	& \textbf{89.84}	& 83.94		 \\

\rowcolor{light-gray} \hspace{1mm} constant-0.3	& \textbf{84.5}/84.83	& \textbf{90.82}	& \textbf{88.31}	& 76.56	& \textbf{92.79}	& \textbf{92.18}	& \textbf{57.24}	& \textbf{89.85}	& \textbf{84.12}		 \\

\rowcolor{light-gray} \hspace{1mm} constant-0.35	& \textbf{84.4}/\textbf{84.99}	& \textbf{90.84}	& \textbf{88.31}	& \textbf{77.81}	& \textbf{92.86}	& \textbf{91.67}	& 55.62	& \textbf{89.88}	& \textbf{84.04}		 \\

\bottomrule
\end{tabular}
\end{sc}
\end{small}
\end{center}
\caption{Downstream performance for different constant schedule configurations. For each model, we report the average accuracy for each task in GLUE. Bold indicates no significant difference from the highest-performing schedule, \emph{P > 0.05}, t-test.}
\label{tab:sweep-constant}
\end{table*}

\begin{table*}[t]
\begin{center}
\begin{small}
\begin{sc}
\begin{tabular}{lccccccccc}
\toprule
Schedule & MNLI-m/mm & QNLI & QQP & RTE & SST-2 & MRPC & CoLA & STS-B & Avg \\
\midrule




\rowcolor{light-gray} \textit{Decreasing} & & & & & & & & & \\

\rowcolor{light-gray} \hspace{1mm} linear-0.3-0.15	& \textbf{84.61}/\textbf{85.13}	& \textbf{90.89}	& \textbf{88.34}	& 76.25	& \textbf{92.71}	& \textbf{91.87}	& \textbf{58.96}	& \textbf{89.87}	& \textbf{84.29}		 \\

\rowcolor{light-gray} \hspace{1mm} linear-0.3-0.2	& \textbf{84.57}/84.89	& \textbf{90.87}	& \textbf{88.33}	& \textbf{77.04}	& \textbf{92.84}	& 91.38	& 57.29	& \textbf{89.78}	& 84.11		 \\

\rowcolor{light-gray} \hspace{1mm} linear-0.3-0.25	& \textbf{84.63}/84.93	& \textbf{90.84}	& \textbf{88.33}	& 76.1	& \textbf{92.84}	& \textbf{92.02}	& 57.33	& \textbf{89.19}	& 84.02		 \\

\midrule
\rowcolor{med-gray} \textit{Increasing} & & & & & & & & & \\

\rowcolor{med-gray} \hspace{1mm} linear-0.3-0.35	& 84.31/84.85	& \textbf{90.73}	& \textbf{88.28}	& \textbf{76.9}	& \textbf{92.91}	& \textbf{91.68}	& 55.85	& 89.7	& 83.91		 \\

\rowcolor{med-gray} \hspace{1mm} linear-0.3-0.4	& 84.19/84.71	& \textbf{90.74}	& \textbf{88.31}	& \textbf{76.82}	& 92.49	& \textbf{91.79}	& 55.67	& 87.83	& 83.62		 \\

\rowcolor{med-gray} \hspace{1mm} linear-0.3-0.45	& 84.07/84.68	& \textbf{90.85}	& \textbf{88.29}	& \textbf{77.02}	& 92.43	& \textbf{91.98}	& 55.84	& \textbf{89.92}	& 83.9		 \\

\bottomrule
\end{tabular}
\end{sc}
\end{small}
\end{center}
\caption{Downstream performance for different linear schedule configurations. For each model, we report the average accuracy for each task in GLUE. Bold indicates no significant difference from the highest-performing schedule, \emph{P > 0.05}, t-test.}
\label{tab:sweep-linear}
\end{table*}

In scheduling the masking rate, we introduce two new parameters: the initial masking rate and the final masking rate.
To determine the optimal configuration of these parameters for the \texttt{BERT-base} experiments, we performed the following search over parameter configurations.
For all experiments, we used the same training setup as presented in Appendix~\ref{sec:appendix-training} and selected the best hyperparameters based on the model's performance on the GLUE benchmark.
We first determined the optimal constant rate, by pretraining with constant masking rates in $\{15\%, 20\%, 25\%, 30\%, 35\% \}$.
After determining that $30\%$ was the optimal masking rate for constant masking schedules (Table~\ref{tab:sweep-constant}), we fixed 30\% to be the starting masking rate for our linear schedules and swept over final masking rates of $\{ 15\%, 20\%, 25\%, 35\%, 40\%, 45\% \}$.
From this sweep, we determined that \texttt{linear-0.3-0.15} was the optimal linear schedule.
Furthermore, decreasing masking rate schedules consistently outperform constant masking rate schedules (Table~\ref{tab:sweep-linear}).

For computational reasons, we did not perform the corresponding sweep over scheduling rates for \texttt{BERT-large}.
Instead, we follow the recommendation of \citet{bert-rate-wettig} and use a 40\% masking rate as the best constant masking rate.
We then propose \texttt{linear-0.4-0.15} as our dynamic schedule following the optimal setting of a 15\% decreasing dynamic schedule observed from our sweep over hyperparameters for \texttt{BERT-base}.

\section{Grammatical Understanding}
\label{sec:appendix-blimp}

\begin{table*}[t]
\rowcolors{3}{light-gray}{med-gray}
\begin{center}
\begin{small}
\begin{sc}
\begin{tabular}{lccc}
\toprule
& \multicolumn{3}{c}{Schedule} \\
\cline{2-4}
\noalign{\smallskip}
Task & linear-0.3-0.15 & constant-0.15 & constant-0.3 \\
\midrule

Anaphor Agreement	& \textbf{98.72}	& \textbf{98.77}	& \textbf{98.63}	\\

Argument Structure	& \textbf{76.13}	& \textbf{76.59}	& 75.36	\\

Binding	& \textbf{76.13}	& \textbf{75.76}	& \textbf{74.91}	\\

Control Raising	& \textbf{78.31}	& \textbf{79.17}	& 77.13	\\

Determiner	& \textbf{95.51}	& \textbf{95.72}	& \textbf{95.43}	\\

Ellipsis	& \textbf{85.38}	& \textbf{84.63}	& \textbf{85.88}	\\

Filler Gap	& \textbf{79.71}	& \textbf{78.37}	& 77.38	\\

Irregular Forms	& \textbf{91.02}	& \textbf{90.0}	& \textbf{90.87}	\\

Island Effects	& \textbf{78.11}	& \textbf{76.17}	& \textbf{78.34}	\\

Npi Licensing	& \textbf{80.62}	& \textbf{80.26}	& \textbf{81.63}	\\

Quantifiers	& \textbf{81.08}	& \textbf{81.79}	& 79.93	\\

Subject Verb Agreement	& \textbf{90.17}	& \textbf{90.37}	& 89.47	\\

Overall	& \textbf{82.7}	& \textbf{82.44}	& 82.13	\\

\bottomrule
\end{tabular}
\end{sc}
\end{small}
\end{center}
\caption{Average accuracy for each super-task in BLiMP. Bold indicates mean + standard error matches best average.}
\label{tab:blimp-subtask}
\end{table*}

In this section, we further detail the BLiMP~\cite{blimp-warstadt} benchmark.

BLiMP sub-tasks are organized into collections of super-tasks that categorize a given linguistic phenomenon.
Each sub-task is composed of minimal pairs of correct (positive) sentences and incorrect (negative) examples.
The model correctly evaluates an example pair if it assigns a higher probability to the positive sentence in the pair than the negative sentence.
However, we note that BERT is not a true language model as it does not produce a probability score over a sequence of tokens.
Accordingly, following \citet{mlm-score-salazar}, we use the \emph{pseudo-log-likelihood (PLL)} to score each sentence.
The PLL is computed by iteratively masking each position in the input sequence and then summing the log likelihood of each masked token.

We present and discuss the average model performance for \texttt{BERT-base} across all tasks in Section \ref{sec:grammar}, finding that \texttt{linear-0.3-0.15} outperforms \texttt{constant-0.3} and has similar performance to \texttt{constant-0.15}.
In table \ref{tab:blimp-subtask}, we present the performance on each individual super-task.
We find that \texttt{linear-0.3-0.15} and \texttt{constant-0.15} have accuracies within one standard error of each other across all super-tasks in BLiMP.
Additionally, \texttt{linear-0.3-0.15} outperforms \texttt{constant-0.3} on 5 out of the 12 BLiMP super-tasks and achieves parity on all other tasks.

\citet{word-order-lasri} found that in a synthetic setting, higher masking rates increase model dependence on positional information and thus improve syntactic understanding.
Interestingly, we find the opposite effect: \texttt{constant-0.15} significantly outperforms \texttt{constant-0.3} on BLiMP.
This observation, combined with the improved overall performance of scheduling, suggests that the improvement in grammar from scheduling is not simply due to being exposed to a higher masking rate.

\section{BERT-Large Downstream Performance Throughout Pretraining}
\label{sec:appendix-large-pareto}

\begin{figure}[ht]
    \begin{center}
        \centerline{\includegraphics[width=\columnwidth]{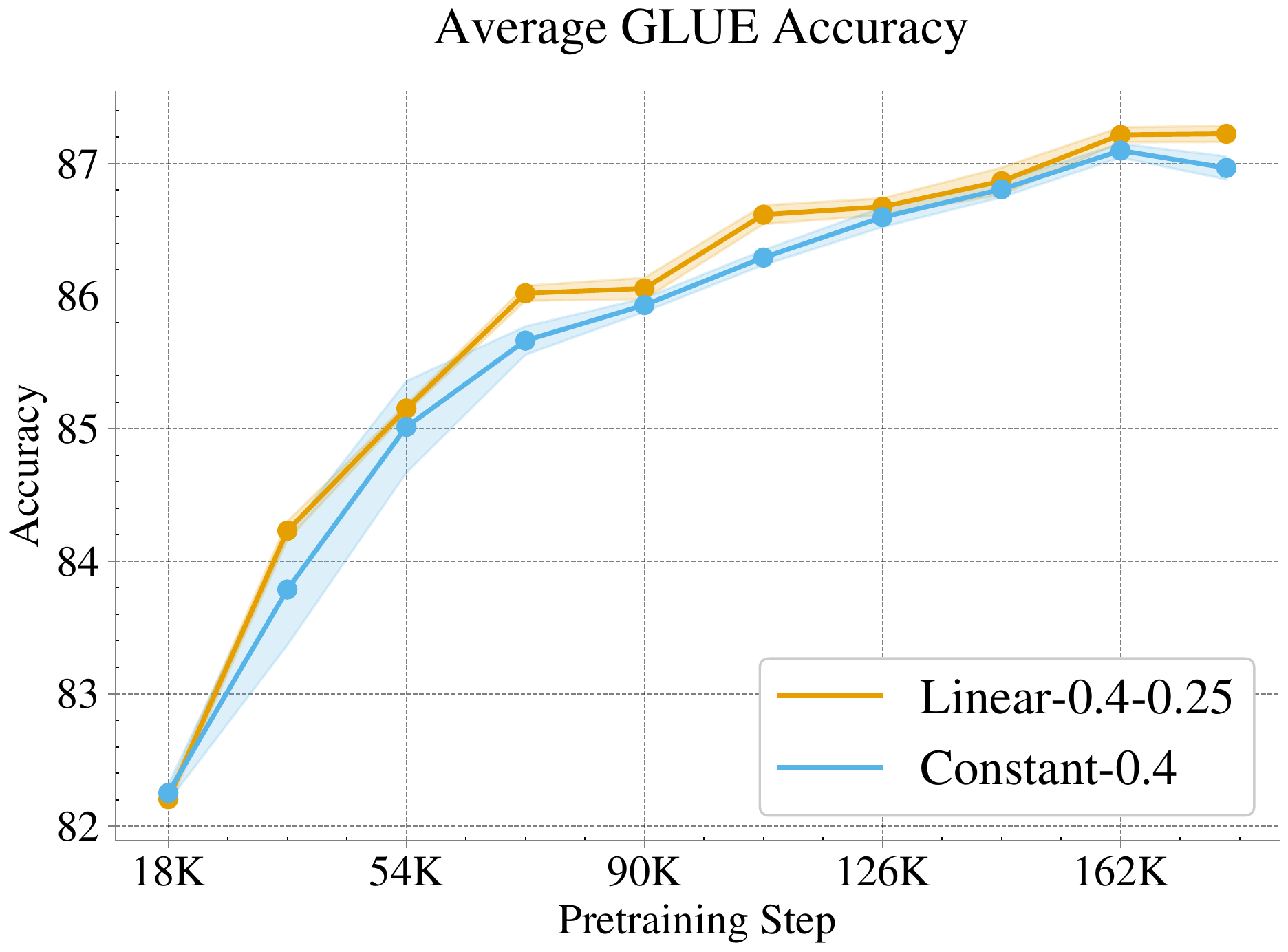}}
        \caption{Average GLUE accuracy evaluated over the course of pretraining for BERT-large.}
        \label{fig:large-speedup}
    \end{center}
\end{figure}

In this section we report the average GLUE performance from different pretraining checkpoints of \texttt{linear-0.4-0.25} and \texttt{constant-0.4} for \texttt{BERT-large} (Figure~\ref{fig:large-speedup}).
We find that \texttt{linear-0.4-0.25} is a Pareto improvement over \texttt{constant-0.4} for each pretraining step evaluated.
This means that \texttt{linear-0.4-0.25} exceeds or matches baseline performance for no increase in training time.

\section{Computing Scheduling Speedup}
\label{sec:appendix-speed-up}

\begin{figure}[ht]
    \begin{center}
        \centerline{\includegraphics[width=\columnwidth]{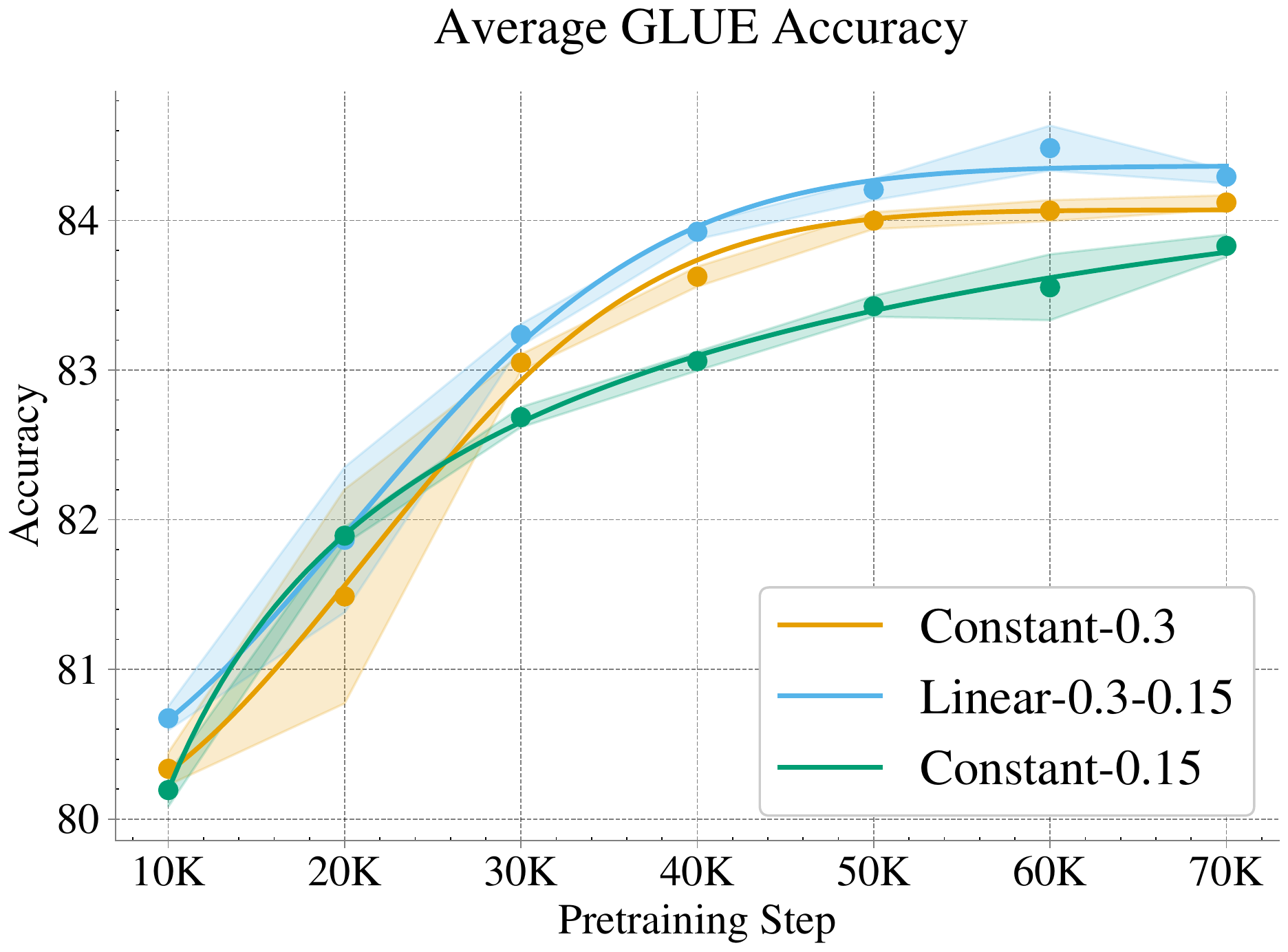}}
        \caption{Pretraining step vs interpolated average GLUE accuracy for \texttt{BERT-base}.}
        \label{fig:regressed-speedup}
    \end{center}
\end{figure}

To compute the efficiency gain of linear scheduling, we evaluate all models on GLUE after every 10K pretraining steps.
We then perform a regression on the number of model steps and the corresponding average GLUE performance using a model of the form:
\begin{equation*}
    c_1 - c_2  \text{exp}\{(-(c_3 t)^{c_4}\}
\end{equation*}
where $c_i$ are the regression variables and $t$ is the pretraining step.
After fitting a model to each schedule's step vs. GLUE performance, we compute the expected speedup by solving for the step in which one schedule achieves the best GLUE performance of the schedule being compared.
We show the regressed pretraining step vs GLUE performance curves in Figure~\ref{fig:regressed-speedup}. We evaluate speedup as a function of pretraining step instead of wall-clock time because dynamic schedules and constant schedules have identical throughput.
\section{Nonlinear Schedules}
\label{sec:appendix-other-sched}

\begin{table*}[t]
\begin{center}
\begin{small}
\begin{sc}
\begin{tabular}{lccccccccc}
\toprule
Schedule & MNLI-m/mm & QNLI & QQP & RTE & SST-2 & MRPC & CoLA & STS-B & Avg \\
\midrule

\rowcolor{light-gray} \textit{Constant} & & & & & & & & & \\

\rowcolor{light-gray} \hspace{1mm} constant-0.15	& 84.3/84.71	& 90.38	& 88.31	& 76.65	& \textbf{92.91}	& \textbf{91.94}	& 55.89	& 89.38	& 83.83		 \\

\rowcolor{light-gray} \hspace{1mm} constant-0.3	& 84.5/84.83	& \textbf{90.82}	& 88.31	& 76.56	& \textbf{92.79}	& \textbf{92.18}	& 57.24	& \textbf{89.85}	& 84.12		 \\

\midrule
\rowcolor{med-gray} \textit{Dynamic} & & & & & & & & & \\

\rowcolor{med-gray} \hspace{1mm} linear-0.3-0.15	& \textbf{84.61}/\textbf{85.13}	& \textbf{90.89}	& \textbf{88.34}	& 76.25	& \textbf{92.71}	& \textbf{91.87}	& \textbf{58.96}	& \textbf{89.87}	& \textbf{84.29}		 \\

\rowcolor{med-gray} \hspace{1mm} cosine-0.3-0.15	& \textbf{84.55}/84.97	& \textbf{90.94}	& \textbf{88.39}	& \textbf{77.67}	& \textbf{92.91}	& \textbf{91.94}	& 57.45	& 89.64	& \textbf{84.27}		 \\

\rowcolor{med-gray} \hspace{1mm} step-0.3-0.15	& \textbf{84.65}/\textbf{85.09}	& \textbf{90.85}	& \textbf{88.37}	& \textbf{77.71}	& \textbf{92.76}	& \textbf{91.56}	& 57.47	& 89.59	& \textbf{84.23}		 \\

\bottomrule
\end{tabular}
\end{sc}
\end{small}
\end{center}
\caption{Downstream performance for different scheduler functions. For each model we report the average accuracy for each task in GLUE.}
\label{tab:other-sched}
\end{table*}

\begin{table*}[t]
\begin{center}
\begin{small}
\begin{sc}
\begin{tabular}{lccccccccc}
\toprule
 Schedule & MNLI-m/mm & QNLI & QQP & RTE & SST-2 & MRPC & CoLA & STS-B & Avg \\
\midrule

rts-constant-0.15	& 83.06/83.46	& 90.64	& 88.22	& \textbf{75.38}	& \textbf{92.06}	& 91.21	& \textbf{56.87}	& 89.92	& 83.42		 \\
rts-constant-0.3	& 83.09/\textbf{83.72}	& \textbf{90.64}	& 88.27	& \textbf{75.74}	& \textbf{91.9}	& 91.15	& 55.41	& 90.02	& 83.33		 \\
rts-linear-0.3-0.15	& \textbf{83.54}/\textbf{83.91}	& \textbf{90.83}	& \textbf{88.37}	& 74.15	& \textbf{92.06}	& \textbf{91.76}	& \textbf{57.53}	& \textbf{90.21}	& \textbf{83.60}		 \\

\bottomrule
\end{tabular}
\end{sc}
\end{small}
\end{center}
\caption{Downstream performance for different random substitution rate schedules. For each model, we report the average accuracy for each task in GLUE. Bold indicates no significant difference from best-performing schedule, \emph{P > 0.05}, t-test.}
\label{tab:rts-best-perf}
\end{table*}

\begin{figure}[ht]
    \begin{center}
        \centerline{\includegraphics[width=\columnwidth]{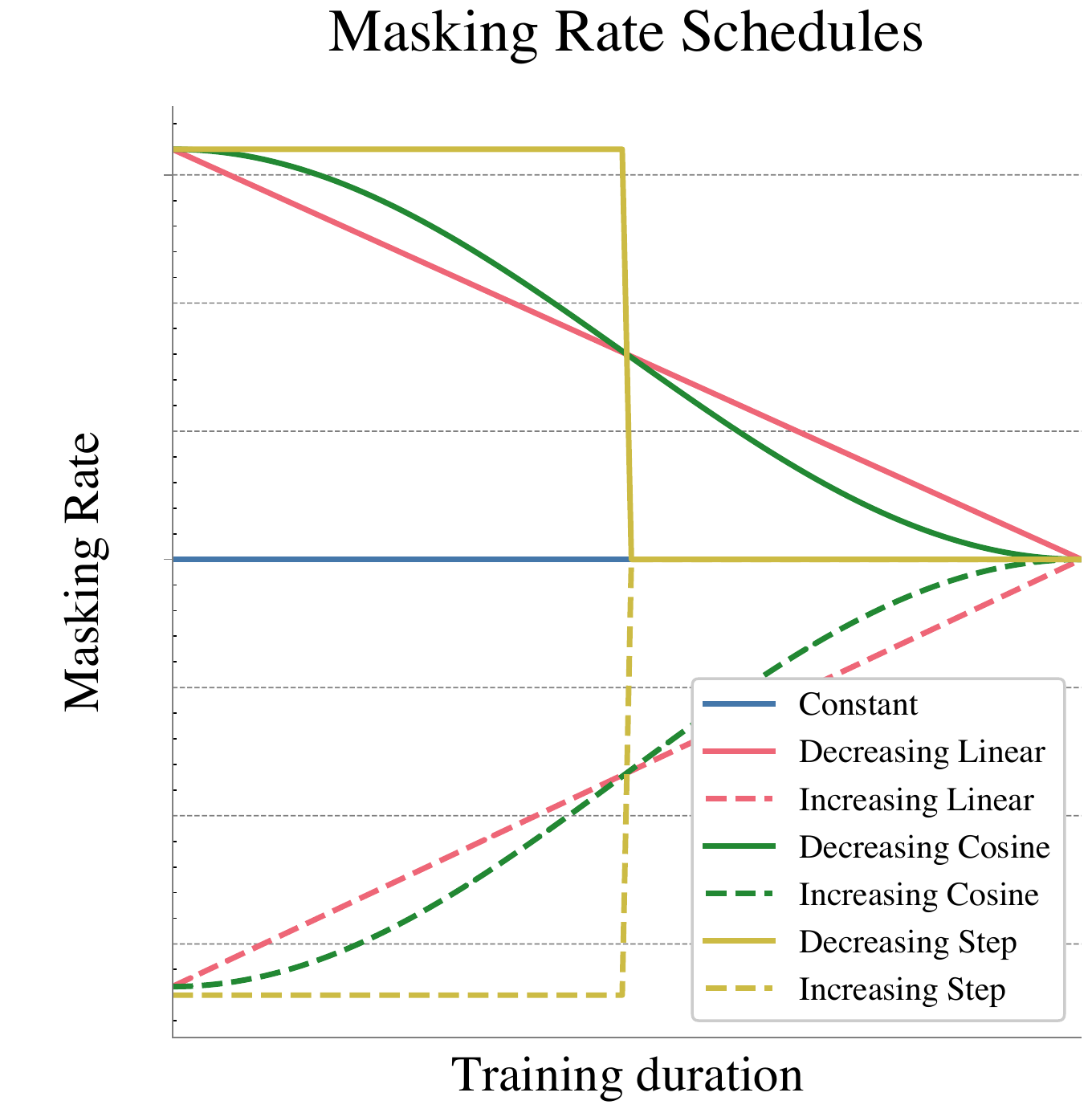}}
        \caption{Various masking rate schedules we considered. Schedules can be constant, increasing or decreasing, and change following a linear, cosine, or step function.}
        \label{fig:schedules}
    \end{center}
\end{figure}

Let $\mathcal{T}_{\texttt{total}}$ be the total number of steps the model takes during training and $t_{i}$ be the current model step.
Let $p_i$ and $p_f$ be the initial and final masking rate respectively.
For each step, we set the masking rate $p_{\text{mask}}$ according to the following schedules.
In Figure~\ref{fig:schedules} we provide a graphical representation of the different schedules experimented with which we detail below.

\paragraph{Cosine scheduling.}
We directly adopt cosine scheduling as proposed in ~\citep{sgdr-loshchilov}.
We perform cosine scheduling by annealing the masking rate following half a cycle of a cosine curve.
The masking rate is then defined as:
\[p_{\text{mask}, t}=p_i + \frac{(p_f - p_i)}{2}*(1 + \cos{((1 -\frac{t}{\mathcal{T}_{\texttt{total}}})\pi)})\]
We refer to cosine schedules as \texttt{cosine-\{$p_i$\}-\{$p_f$\}}.

\paragraph{Step-wise scheduling.}
Step wise scheduling is defined by a decay rate, $\gamma$, and a set of timesteps, $\Gamma = \{t_1,...,t_m\}$, for when the masking rate is decayed.
The schedule is then defined as:
\[ p_{\text{mask}, t}=
\begin{cases}
    \gamma*p_{\text{mask}, t-1}, & t \in \Gamma \\
    p_{\text{mask}, t-1}
\end{cases}
\]
Our experiments are restricted to step-wise schedules that apply the decay to the masking rate only once, halfway through the training duration.
As such, for ease of notation, we ignore the decay rate when talking about step-wise schedules and instead describe our step-wise schedules in terms of their initial and final masking rates.
We refer to step-wise schedules as \texttt{step-\{$p_i$\}-\{$p_f$\}}.

\subsection{Results}

Following the same pretraining and evaluation setup (Section \ref{sec:appendix-training}), we evaluate the performance of \texttt{cosine-0.3-0.15} and \texttt{step-0.3-0.15}.
We find that for linear, cosine, and step-wise scheduling there is no statistically significant difference in average GLUE performance (Table~\ref{tab:other-sched}).
We find that \texttt{linear-0.3-0.15} outperforms \texttt{cosine-0.3-0.15} on 3 tasks, underperforms on 1 task, and achieves parity on the rest of the tasks in GLUE.
Similarly, \texttt{linear-0.3-0.15} outperforms \texttt{step-0.3-0.15} on 2 tasks, underperforms on 1 task, and achieves parity on the rest of the tasks in GLUE.
In the context of these results, we conclude that the scheduler type is less significant than the schedule parameters, and as such conduct the primary experiments in our paper with respect to the simple linear scheduler.

\section{Generalization to Other Objectives}
\label{sec:appendix-rts}
\subsection{Set-Up}
In order to further demonstrate the success of dynamically scheduling the pretraining objective for encoder transformers, we evaluate dynamically scheduling the token substitution in the Random Token Substitution (RTS) objective \cite{rts-liello}.
In the RTS objective a subset of tokens, defined by the random token substitution rate, are randomly substituted with another token in the vocabulary.
The model is then trained to classify whether a token was randomly substituted or is the original token.
The random token substitution rate was originally set to be a constant $15\%$.
In our work, we experiment both with a constant $30\%$ and linearly decreased from $30\%$ to $15\%$ random token substitution rate.

All other hyperparameters and data choices are the same as the ones we used for MLM training of \texttt{BERT-base} (Appendix~\ref{sec:appendix-training}).

\subsection{Results}

\paragraph{Improvement in final performance}
We examine the effect of scheduling the random token substitution rate on downstream GLUE performance (Table~\ref{tab:rts-best-perf}).
As \texttt{rts-constant-0.15} is the better-performing constant schedule for RTS, we focus our comparison on this baseline.
We find that \texttt{rts-linear-0.3-0.15} outperforms \texttt{rts-constant-0.15} on 6 out of the 8 tasks in GLUE, and only performs worse on 1 task, leading to an average improvement on GLUE of $0.18\%$.
This result demonstrates that the improved gains from dynamically scheduling the pretraining objective for \texttt{BERT} style models also generalize to the RTS task.
\\
\paragraph{Performance throughout pretraining}

\begin{figure}[ht]
    \begin{center}
        \centerline{\includegraphics[width=\columnwidth]{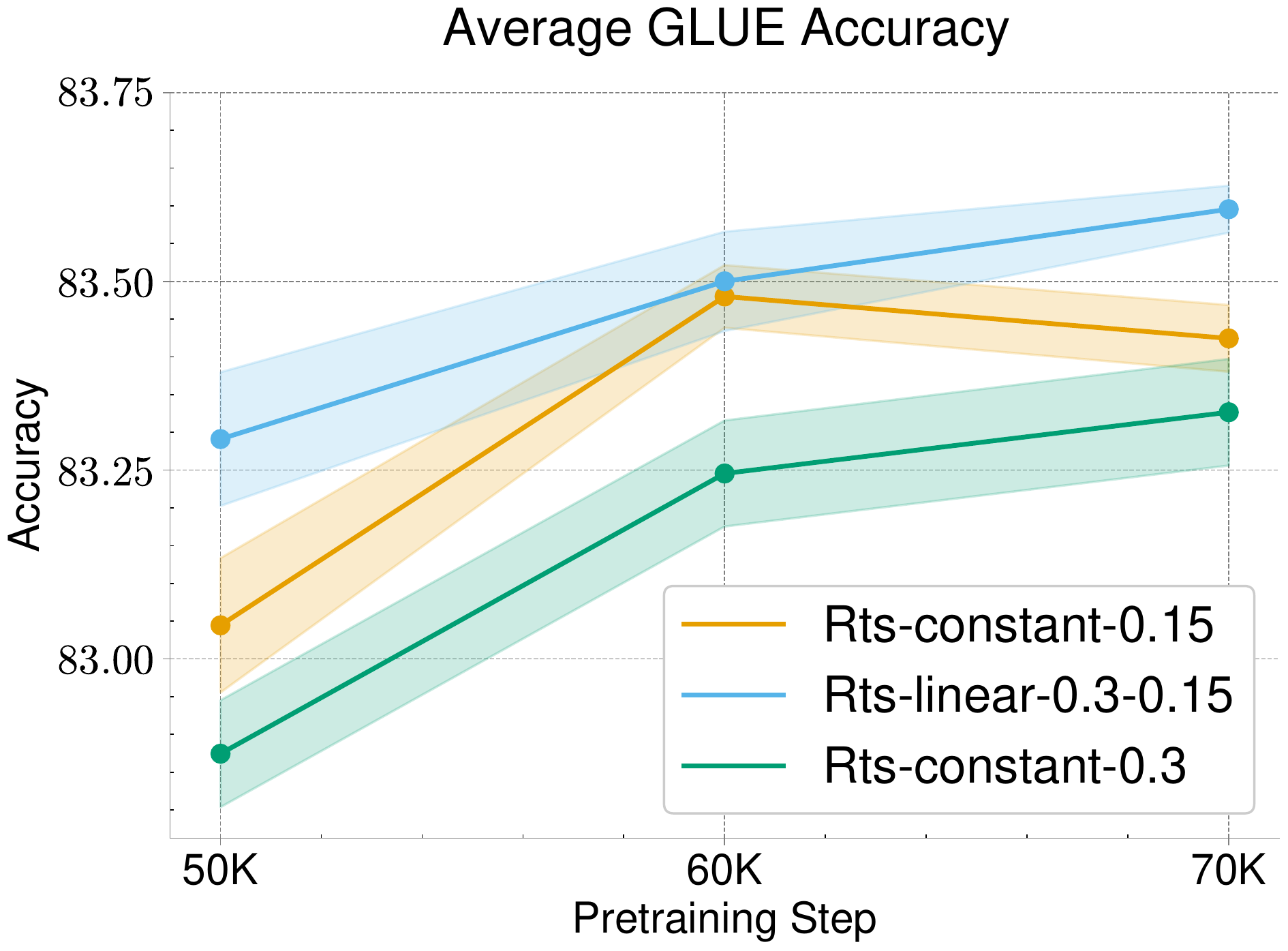}}
        \caption{Pretraining step vs interpolated average GLUE accuracy for RTS with \texttt{BERT-base}.}
        \label{fig:rts-speedup}
    \end{center}
\end{figure}

We examine the effect at different points of pre-training of scheduling the random token substitution rate.
Specifically, we compute the downstream GLUE accuracy for the different schedules at 50K, 60K, and 70K of training.
We find that \texttt{rts-linear-0.3-0.15} is a Pareto improvement over both \texttt{rts-constant-0.3} and \texttt{rts-constant-0.15}, meaning linear scheduling performs better for each intermediate checkpoint evaluated (Figure~\ref{fig:rts-speedup}).

\end{document}